\documentclass[runningheads, dvipdfmx]{llncs}
\usepackage{graphicx}
\usepackage{amsmath}

\begin{document}
\title{Development of Quantized DNN Library for Exact Hardware Emulation}
\author{Masato Kiyama\inst{1} \and
Motoki Amagasaki\inst{1} \and
Masahiro Iida\inst{1}}
    
\authorrunning{M. Kiyama et al.}
\institute{Faculty of Advanced Science and Technology, Kumamoto University 2-39-1 Kurokami, Chuo-ku, Kumamoto 860-8555, Japan \\
\email{\{masato, amagasaki, iida\}@cs.kumamoto-u.ac.jp}}

\maketitle

\begin{abstract}
Quantization is used to speed up execution time and save power when runnning Deep neural networks (DNNs) on edge devices like AI chips.
To investigate the effect of quantization,
we need performing inference after quantizing the weights of DNN with 32-bit floating-point precision by a some bit width, and then quantizing them back to 32-bit floating-point precision.
This is because the DNN library can only handle floating-point numbers.
However, the accuracy of the emulation does not provide accurate precision.
We need accurate precision to detect overflow in MAC operations or to verify the operation on edge devices.
We have developed PyParch, a DNN library that executes quantized DNNs (QNNs) with exactly the same behavior as hardware.
In this paper, we describe a new proposal and implementation of PyParch.
As a result of the evaluation, the accuracy of QNNs with arbitrary bit widths can be estimated for large and complex DNNs such as YOLOv5, and the overflow can be detected.
We evaluated the overhead of the emulation time and found that it was 5.6 times slower for QNN and 42 times slower for QNN with overflow detection compared to the normal DNN execution time.
\end{abstract}

\section{Introduction}

DNNs (Deep Neural Networks) have shown outstanding performance in many areas such as image classification and object detection\cite{DNN}.
DNN is a class of machine-learning algorithms based on artificial neural networks (ANNs), which are a computational model inspired by the human brain and how it perceives information through the interaction of neurons.
DNNs have three types of layers: an input layer, hidden layers, and an output layer.
There are weights between the layers that indicate the strength of the connections between the artificial neurons.
In DNNs, each edge has a unique weight and each node has a unique bias. An accuracy of a neural net depends on its weights and biases after training.

DNNs have already surpassed human capabilities in certain areas\cite{AlphaGo,BERT}.
Those DNNs require a large amount of weights.
As a result, the amount of compute used in training and inference grow, and the processing time becomes long.
In order to reduce this problem, special hardware such as GPU (Graphic Processor Unit) is good candidate.
However, these specific hardwares are not always available on edge devices like mobile gadget.
Then, it is difficult to run large-scale DNNs on mobile devices.
In order to use DNNs on mobile devices, a number of methods have been developed\cite{Compression,LOW}.

Quantization is a effective technique to reduce the bit-size of weights in DNNs.
In DNNs, the output is computed from the input and the weights, and inference is performed using floating point (FP32) with 32-bit precision.
In quantization, the values are represented with a smaller bit width for inference.
Quantization uses fewer bits to represent the values and make inference lightweight, thus speeding up the computation and reducing memory usage.

In this paper, we present a new proposal and implementation based on PyParch\cite{PyParch0,PyParch1}.
PyParch runs on PyTorch\cite{PYTORCH}, which makes it easy to extend.
In addition, the emulation results of DNNs quantized to fixed-point arbitrary bits are the same as those of hardware execution.
Therefore, the optimal bit width (arithmetic precision) for hardware implementation of the quantized model can be accurately estimated.

The remainder of this paper is structured as follows.
Section 2 describes related work in quantization.
Section 3 describes the software emulation with quantization and proposes our hardware calculation method.
Section 4 evaluates our method.
Finally, Section 5 summarizes and concludes the paper.

\section{Related Work}

Quantization is a technique to compress and accelerate DNN inference.
Neural network quantization has become an important research area due to its great impact on the deployment of large models to resource-constrained devices, s
uch as AI chips.
Quantization greatly reduces the amount of weights and activation data.

Quantization can be classified into two type: linear quantization and non-linear quantization \cite{FPGAsurvey}.
Linear quantization finds the nearest fixed-point representation.
Non-linear quantization independently assigns values to different binary codes.
In this section, we focus on linear quantization method.

Tensorflow-Lite\cite{TFLite} converts a trained model with FP32 weights to 8-bit (INT8) integers, performs inference, and stores the results in FP32.
QNNPACK\cite{QNNPACK} handles data in INT8 fixed-point format.
The following libraries emulate quantized DNNs: TensorQuant\cite{TensorQuant}, PyTorch\cite{PYTORCH}, Tensorflow\cite{Tensorflow}, and QPyTorch\cite{QPyTorch}.
QPyTorch is similar to our method. Their method runs on PyTorch and supports many numerical expressions and rounding operations.
However, They do not support the layer fusion and overflow detection proposed in our method.
Our method differs from other quantization libraries in that it combines fixed-point quantization with arbitrary bit-width, quantization-specific layer merging, and overflow detection.

\section{Implementation}
In this section, we describe our new implementation of PyParch.
An emulation process of a quantized model on Pyparch is as follows.

\begin{enumerate}

  \item Fusion of layers
  \item Quantize weights for each layer
  \item Calibration
  \item Replace the layer that contains the division
  \item Adjust the FL(Fraction Length) in the combination and addition
  \item Quantize outputs of a layer

\end{enumerate}

Section \ref{sec:quantization} describes the FP32 to fixed-point conversion method.
In Section \ref{sec:fuse}, we describe the layer fusion that are not supported by other libraries and how to solve them.
Section \ref{sec:replace} discusses how to replace layers containing division.
Section \ref{sec:adjust} discusses how to maintain proper FL.
Section \ref{sec:overflow} describes our overflow detection method.

\subsection{Quantization}
\label{sec:quantization}

The conversion method for quantizing FP32 to fixed-point uses the formula (\ref{eq:quant}) and the conversion method for inverse quantization from fixed-point to FP32 uses the formula (\ref{eq:dequant}).
WL (Word Length) is the fixed-point bit width and FL (Fraction Length) is the width of the decimal point.
The definition of clamp is given by the expression (\ref{eq:clamp}), and the rounding process is round-half-away-from-zero.

\begin{equation}\label{eq:quant}
  x_{INT} = clamp\left(round\left(x_{FP32} \times 2^{FL}\right), -2^{WL - 1}, 2^{WL - 1} - 1\right)
\end{equation}

\begin{equation}\label{eq:dequant}
  x_{FP32} = x_{INT} \times 2^{-FL}
\end{equation}

\begin{equation}\label{eq:clamp}
  clamp(x, min, max) = \begin{cases}
    min & x \leq min \\
    x & a < x < b \\
    max & x \geq max
    \end{cases}
\end{equation}

Previous method\cite{PyParch1} use INT that are quantized from FP32 in inference for exact emulation.
On the other hand, the present method use quantized FP32 that are dequantized from INT.
This is because if WL is set appropriately and all the division performed in the layer is converted to multiplication and emulated, the result is the same as the inference using INT.
Other quantization libraries don't consider this point, so the emulation results do not match the inference results on hardware.
This is because INT is not closed to division, and the correspondence between quantized FP32 and INT is broken.
In the case of addition and multiplication, if the bit width is enough, the correspondence is not broken because INT is closed for each operation, and accurate emulation can be performed.

In our emulation, we assume that all hardware operations are performed by INT with no division circuit.
The hardware cost of a divider is high, so many AI chips have only addition, subtraction, and multiplication.
It is necessary to add a separate dedicated calculation for devices with divider.

\subsection{Layer Fusion}
\label{sec:fuse}

\begin{figure}[t]
\begin{center}
\includegraphics[scale=0.2]{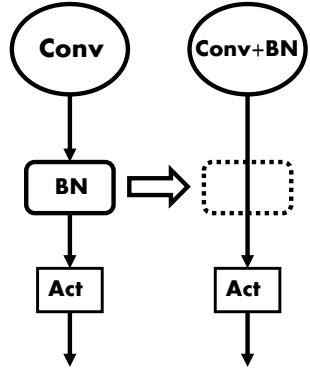}
\end{center}
\caption{layer fusion}
\label{fig:fuse}
\end{figure}

Layer fusion is one of the optimization step during inference.
In our method, we fuse the convolutional layer and the batch normalization layer.
Fig.~\ref{fig:fuse} shows the change in structure before and after layer fusion.
This fusion is used to reduce the computational complexity.
In PyParch, this step not only reduces the number of operations, but also eliminates the division of the batch normalization layer for exact emulation.

\begin{figure*}[t]
\begin{center}
\includegraphics[scale=0.2]{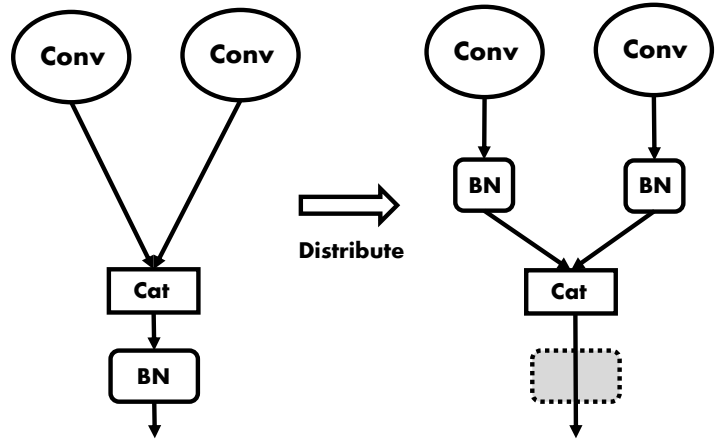}
\end{center}
\caption{distribution}
\label{fig:distribute}
\end{figure*}

This fusion which is available in other libraries can be fused only under the condition for batch normalization layer next to the convolutional layer.
We use the object detection model YOLOv5\cite{YOLOV5} to check the effectiveness of our method.
Because the batch normalization layer is performed after combining the output results of the two convolutional layers in YOLOv5, layer fusion cannot be performed in other libraries.
Therefore, in our method, the batch normalization layer after the merging is split and distributed to each corresponding convolutional layer.
Adapting this method changes the structure of the model for batch normalization layer after the convolutional layer.
Next, the convolutional layer and the batch normalization layer can be merged.
Fig.~\ref{fig:distribute} shows the change in structure before and after the distribution.

\subsection{Layer Replacement}
\label{sec:replace}

As mentioned in the \ref{sec:quantization} section, exact emulation of quantized DNNs is not possible if layer operations include division.
The convolutional and fully connected layer does not include division, but the Global Average Pooling (GAP) \cite{GAP} and the activation function hard-swish \cite{HSWISH} contain division in their layers.
In our method, we use PyTorch's capabilities to detect layers and activation functions that include division and replace them to not include division.

LeakyReLU\cite{Leaky} is an activation function that does not include division, but cannot be exactly emulated without considering the bit width.
In LeakyReLU, negative numbers are multiplied by a constant. Therefore, it is necessary to quantize the constant with an appropriate bit width.
In our method, we use the PyTorch function to detect LeakyReLU, quantize the constant to be multiplied by the negative number to an appropriate bit width, and replace the layer.

\subsection{Adjustment of FL}
\label{sec:adjust}

\begin{figure}[t]
\begin{center}
\includegraphics[scale=0.25]{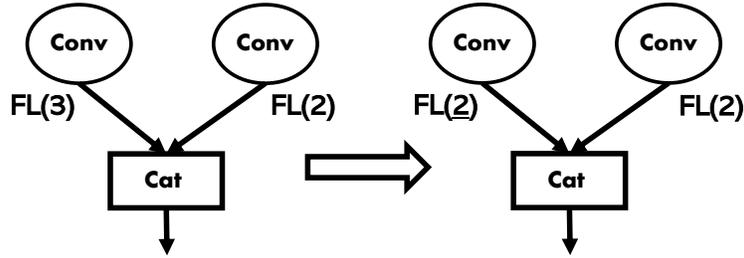}
\end{center}
\caption{Adjustment of Fl}
\label{fig:adjust}
\end{figure}

In our method, each FL is different for each layer to account for the variability in the distribution of weights and output values for each layer.
In a simple DNN model, the output FL of the previous layer becomes the input FL of the layer.
However, if the outputs from multiple layers are added and combined, as in ResNet\cite{Resnet}, the output FL of the layers needs an adjustment.

PyTorch can detect layers easily, but the addition and concatination of the output are difficult to detect becase these are not a layer in PyTorch.
If we convert a DNN model defined in PyTorch into a ONNX\cite{ONNX} format, we can detect the addition and concatination becase these are a node in ONNX.
However, We have to consider the correspondence between a addition and concatination in PyTorch and a node in ONNX.

Therefore, Our method uses PyTorch's ability to calculate the gradient.
Our method trace the gradient information from the output data to identify the points where addition and concatination and adjust the output FL.

If the operation is a concatination, it adjusts the output FL of all layers to the minimum value of the output FL in several previous layers.
Fig.~\ref{fig:adjust} shows the changes before and after the adjustment.
If the operation is an addition, there are two possible methods.
The first is the same method as for concatination.
The second is to change the bit widths to match the maximum values of the previous layers, add them, and then change the bit widths back to the original values for the input of the next layer.
This method does not require any special processing on the software side, but requires control on the hardware side to maintain the appropriate bit width.
Compared to the first method, the quantized DNN may be more accurate because it retains more information.
Of course, a target hardware has to require a mechanism to realize the first method.

\subsection{Overflow Detection}
\label{sec:overflow}

When designing hardware, it is necessary to have information on the number of bits required for overflow during multiply-accumulate (MAC) operations.
In our method, an arbitrary number of carry bits can be set in the emulation for overflow detection.
The previous version of PyParch checked overflow using bit manipulation after MAC operations.
it took a lot of emulation time and was not feasible for large DNN models.
In out method, we use PyTorch's im2col\cite{IM2COL} to convert the input and weights of the convolutional layer, and pass the data to a function in an extended library written in Rust.

\section{Experiments}

\begin{figure}[t]
\begin{center}
\includegraphics[scale=0.25]{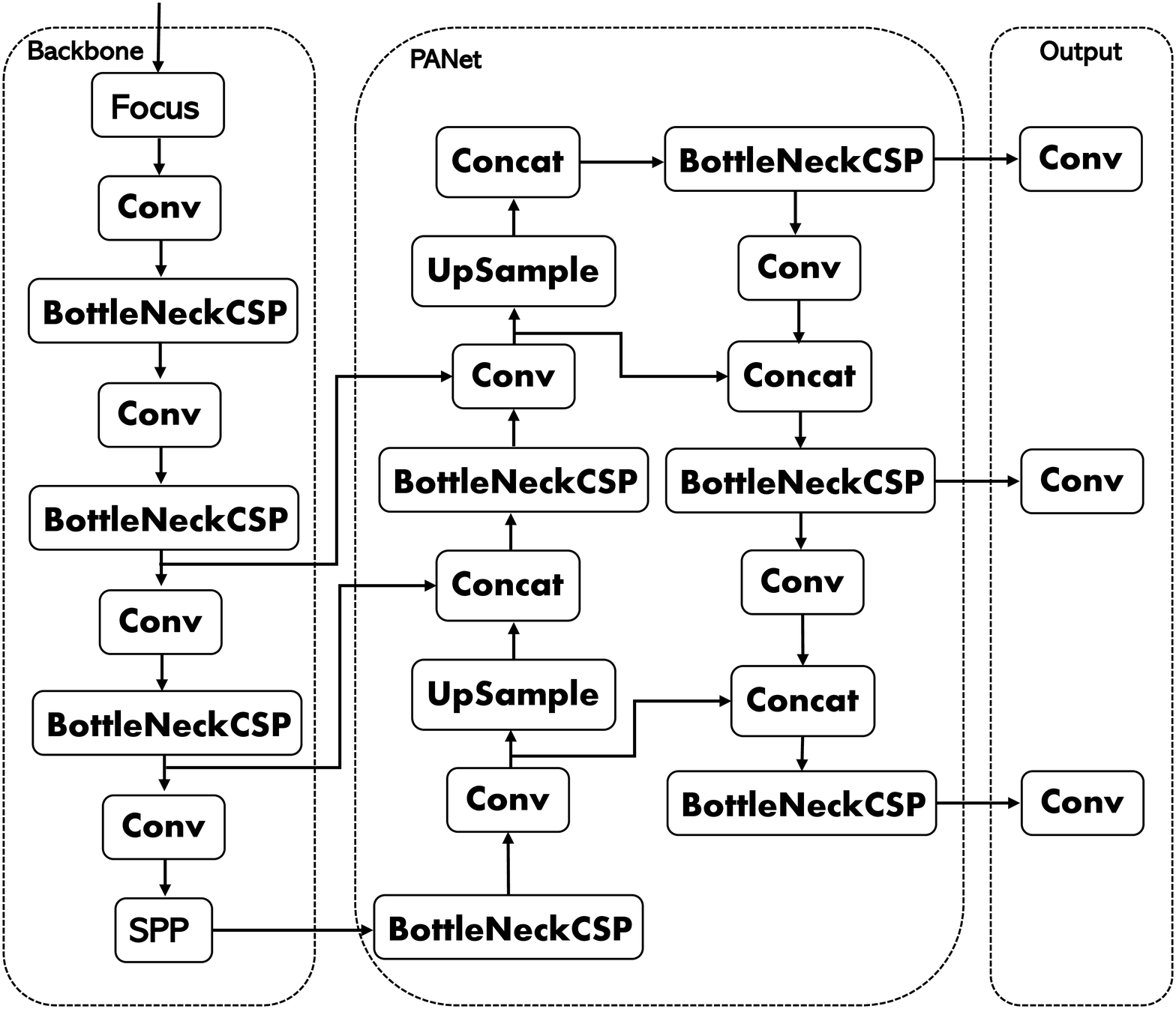}
\end{center}
\label{fig:yolov5}
\caption{Overview of YOLOv5}
\end{figure}

\subsection{Datasets and DNN Models}\label{result:method}

Our experiments are performed on object detection model YOLOv5\cite{YOLOV5} to verify the effectiveness of our method.
YOLOv5 model is shown in Fig.~\ref{fig:yolov5} and the evaluation environment is shown in Table~\ref{table:4}.

YOLOv5 has several models depending on accuracy and speed.
We used the lowest model weight.
The dataset used is the 128-image dataset used in test.py, which is available on the YOLOv5 github\footnote{https://github.com/ultralytics/yolov5}.
The trained model to be quantized was the data made available by the authors of YOLOv5.
YOLOv5 for evaluation is the model trained in COCO 2017\cite{COCO}, and the class to be recognized is 80.

\begin{table}[tb]
\caption{evaluation environment}
\label{table:4}
\begin{center}
\begin{tabular}{l|l}
\hline
Name & Configuration \\
\hline
CPU & Intel(R) Core(TM) 2.8GHz \\
Memory & 16GB \\
PyTorch & 1.70 \\
Rust & 1.49.0 \\
\hline
\end{tabular}%
\end{center}
\end{table}

\subsection{Experimental Results}\label{result}

The mAP (mean Average Precision) is used as the evaluation index.
IoU (Intersection over Union) thresholds of 0.5 and 0.95 are used as mAP.5 and mAp.95, respectively.
The results of mAP with changing the WL are shown in Table~\ref{table:1}.
In the original unquantized model, mAP.5 is 0.7 and mAp.95 is 0.459.
WLs of 8 or less are too small to evaluate, and WLs of 17 or more are not evaluated because no degradations are observed.
From the results in Table~\ref{table:1}, YOLOv5 performs as well as the original accuracy when WL is 12.

\begin{table}[tb]
\caption{mAP}
\label{table:1}
\begin{center}
\begin{tabular}{c|l|l}
\hline
WL & mAP.5 & mAP.95 \\
\hline
9 & 0.145 & 0.075\\
10 & 0.435 & 0.245\\
11 & 0.642 & 0.397\\
12 & 0.699 & 0.445\\
13 & 0.708 & 0.458\\
14 & 0.712 & 0.454\\
15 & 0.705 & 0.456\\
16 & 0.714 & 0.457\\
\hline
\end{tabular}%
\end{center}
\end{table}

Each layer has a different distribution of its weights, and quantization is performed accordingly.
Since the output of the activation function has a larger distribution bias than that of the convolutional layer, there is a method that changes the WL of each of the quantization layers \cite{DoReFa}.
In our method, we also increased the WL only for the internal operation of the activation function.
WL was set to 10 and FL was increased to 3.
The results are shown in Table~\ref{table:2}.
From the results in Table~\ref{table:2}, the accuracy of the model improved when WL was increased only for the activation function.
Since we only need to increase WL when computing the activation function, we can expect a reduction in hardware cost.

\begin{table}[tb]
\caption{FL}
\label{table:2}
\begin{center}
\begin{tabular}{c|l|l}
\hline
WL(FL) & mAP.5 & mAP.95 \\
\hline
10(0) & 0.435 & 0.245\\
11(1) & 0.618 & 0.372\\
12(2) & 0.637 & 0.396\\
13(3) & 0.65 & 0.401\\
\hline
\end{tabular}%
\end{center}
\end{table}

The results of the emulation time are shown in Table~\ref{table:3}.
The base is executed without quantization.
Pyparch is the result of quantization and overflow is the result of the method described in \ref{sec:overflow} with the addition of overflow detection.
When executing with quantization, the outputs of the layer are re-quantized to maintain the appropriate bit width.
Therefore, emulation time was 5.6 times slower than the base.
In addition, emulation time with overflow detection was almost 42 times slower than the base.
For further speedup, it is necessary to implement a new layer that supports quantization.

\begin{table}[tb]
\caption{emulation overhead}
\label{table:3}
\begin{center}
\begin{tabular}{c|l}
\hline
model & sec \\
\hline
base & 30.1 \\
pyparh & 167.3 \\
overflow & 1263.3 \\
\hline
\end{tabular}%
\end{center}
\end{table}

\section{Conclusion}

In this paper, we proposed a new method on PyParch and evaluated it.
We proposed and implemented a new method of layer fusion and detection of the overflow for large DNNs.
As a result of evaluation, we were able to estimate the accuracy of quantized DNNs with arbitrary bit width and to detect overflow.
The emulation time was 5.6 times slower for the quantized DNN and 42 times slower for the quantized DNN with overflow detection.

Future research should improve the emulation speed and to automatically convert the quantized model into a format that can be executed on FPGAs.

\bibliographystyle{splncs04}
\bibliography{paper}

\begin{thebibliography}{10}
\providecommand{\url}[1]{\texttt{#1}}
\providecommand{\urlprefix}{URL }
\providecommand{\doi}[1]{https://doi.org/#1}

\bibitem{Tensorflow}
Abadi, M., Agarwal, A., Barham, P., Brevdo, E., Chen, Z., Citro, C., Corrado,
  G.S., Davis, A., Dean, J., Devin, M., Ghemawat, S., Goodfellow, I., Harp, A.,
  Irving, G., Isard, M., Jia, Y., Jozefowicz, R., Kaiser, L., Kudlur, M.,
  Levenberg, J., Man\'{e}, D., Monga, R., Moore, S., Murray, D., Olah, C.,
  Schuster, M., Shlens, J., Steiner, B., Sutskever, I., Talwar, K., Tucker, P.,
  Vanhoucke, V., Vasudevan, V., Vi\'{e}gas, F., Vinyals, O., Warden, P.,
  Wattenberg, M., Wicke, M., Yu, Y., Zheng, X.: {TensorFlow}: Large-scale
  machine learning on heterogeneous systems (2015),
  \url{https://www.tensorflow.org/}, software available from tensorflow.org

\bibitem{ONNX}
Bai, J., Lu, F., Zhang, K., et~al.: Onnx: Open neural network exchange.
  https://github.com/onnx/onnx (2019)

\bibitem{IM2COL}
Chellapilla, K., Puri, S., Simard, P.: {High Performance Convolutional Neural
  Networks for Document Processing}. In: Lorette, G. (ed.) {Tenth International
  Workshop on Frontiers in Handwriting Recognition}. {Universit{\'e} de Rennes
  1}, {Suvisoft}, La Baule (France) (Oct 2006),
  \url{https://hal.inria.fr/inria-00112631}, http://www.suvisoft.com

\bibitem{Compression}
Cheng, Y., Wang, D., Zhou, P., Zhang, T.: A survey of model compression and
  acceleration for deep neural networks. CoRR  \textbf{abs/1710.09282} (2017)

\bibitem{BERT}
Devlin, J., Chang, M.W., Lee, K., Toutanova, K.: {BERT}: Pre-training of deep
  bidirectional transformers for language understanding. In: Proceedings of the
  2019 Conference of the North {A}merican Chapter of the Association for
  Computational Linguistics: Human Language Technologies, Volume 1 (Long and
  Short Papers). pp. 4171--4186. Association for Computational Linguistics,
  Minneapolis, Minnesota (Jun 2019). \doi{10.18653/v1/N19-1423},
  \url{https://www.aclweb.org/anthology/N19-1423}

\bibitem{QNNPACK}
Dukhan, M., Wu, Y., Lu, H.: Qnnpack: open source library for optimized mobile
  deep learning, \url{https://github.com/pytorch/QNNPACK}

\bibitem{FPGAsurvey}
Guo, K., Zeng, S., Yu, J., Wang, Y., Yang, H.: [dl] a survey of fpga-based
  neural network inference accelerators. ACM Trans. Reconfigurable Technol.
  Syst.  \textbf{12}(1) (Mar 2019). \doi{10.1145/3289185},
  \url{https://doi.org/10.1145/3289185}

\bibitem{LOW}
Han, Q., Hu, Y., Yu, F., Yang, H., Liu, B., Hu, P., Gong, R., Wang, Y., Wang,
  R., Luan, Z., Qian, D.: Extremely low-bit convolution optimization for
  quantized neural network on modern computer architectures. In: 49th
  International Conference on Parallel Processing - ICPP. ICPP '20, Association
  for Computing Machinery, New York, NY, USA (2020).
  \doi{10.1145/3404397.3404407}, \url{https://doi.org/10.1145/3404397.3404407}

\bibitem{Resnet}
{He}, K., {Zhang}, X., {Ren}, S., {Sun}, J.: Deep residual learning for image
  recognition. In: 2016 IEEE Conference on Computer Vision and Pattern
  Recognition (CVPR). pp. 770--778 (2016). \doi{10.1109/CVPR.2016.90}

\bibitem{HSWISH}
Howard, A., Pang, R., Adam, H., Le, Q.V., Sandler, M., Chen, B., Wang, W.,
  Chen, L., Tan, M., Chu, G., Vasudevan, V., Zhu, Y.: Searching for
  mobilenetv3. In: 2019 {IEEE/CVF} International Conference on Computer Vision,
  {ICCV} 2019, Seoul, Korea (South), October 27 - November 2, 2019. pp.
  1314--1324. {IEEE} (2019). \doi{10.1109/ICCV.2019.00140},
  \url{https://doi.org/10.1109/ICCV.2019.00140}

\bibitem{TFLite}
Jacob, B., Kligys, S., Chen, B., Zhu, M., Tang, M., Howard, A., Adam, H.,
  Kalenichenko, D.: Quantization and training of neural networks for efficient
  integer-arithmetic-only inference. In: Proceedings of the IEEE Conference on
  Computer Vision and Pattern Recognition (CVPR) (June 2018)

\bibitem{YOLOV5}
Jocher, G., Stoken, A., Borovec, J., NanoCode012, ChristopherSTAN, Changyu, L.,
  Laughing, Hogan, A., lorenzomammana, tkianai, yxNONG, AlexWang1900, Diaconu,
  L., Marc, wanghaoyang0106, ml5ah, Doug, Hatovix, Poznanski, J., Yu, L.,
  changyu98, Rai, P., Ferriday, R., Sullivan, T., Xinyu, W., YuriRibeiro,
  Claramunt, E.R., hopesala, pritul dave, yzchen: ultralytics/yolov5: v3.0 (Aug
  2020). \doi{10.5281/zenodo.3983579},
  \url{https://doi.org/10.5281/zenodo.3983579}

\bibitem{PyParch1}
{Kiyama}, M., {Nakahara}, Y., {Amagasaki}, M., {Iida}, M.: A quantized neural
  network library for proper implementation of hardware emulation. In: 2019
  Seventh International Symposium on Computing and Networking Workshops
  (CANDARW). pp. 136--140 (2019). \doi{10.1109/CANDARW.2019.00032}

\bibitem{PyParch0}
Kiyama, M., Amagasaki, M., Iida, M.: Deep learning framework with arbitrary
  numerical precision. 13th IEEE International Symposium on Embedded
  Multicore/Many-core Systems-on-Chip, MCSoC 2019, Singapore, Singapore,
  October 1-4, 2019 pp. 81--86 (2019). \doi{10.1109/MCSoC.2019.00019}

\bibitem{DNN}
LeCun, Y., Bengio, Y., Hinton, G.: Deep learning. Nature  \textbf{521}(7553),
  436--444 (2015). \doi{10.1038/nature14539},
  \url{https://doi.org/10.1038/nature14539}

\bibitem{GAP}
Lin, M., Chen, Q., Yan, S.: Network in network (2013),
  \url{http://arxiv.org/abs/1312.4400}, cite arxiv:1312.4400Comment: 10 pages,
  4 figures, for iclr2014

\bibitem{COCO}
Lin, T., Maire, M., Belongie, S.J., Bourdev, L.D., Girshick, R.B., Hays, J.,
  Perona, P., Ramanan, D., Doll{'{a} }r, P., Zitnick, C.L.: Microsoft {COCO:}
  common objects in context. CoRR  \textbf{abs/1405.0312} (2014),
  \url{http://arxiv.org/abs/1405.0312}

\bibitem{TensorQuant}
Loroch, D.M., Pfreundt, F.J., Wehn, N., Keuper, J.: Tensorquant: A simulation
  toolbox for deep neural network quantization. In: Proceedings of the Machine
  Learning on HPC Environments, pp.~1--8 (2017)

\bibitem{PYTORCH}
Paszke, A., Gross, S., Massa, F., Lerer, A., Bradbury, J., Chanan, G., Killeen,
  T., Lin, Z., Gimelshein, N., Antiga, L., Desmaison, A., Kopf, A., Yang, E.,
  DeVito, Z., Raison, M., Tejani, A., Chilamkurthy, S., Steiner, B., Fang, L.,
  Bai, J., Chintala, S.: Pytorch: An imperative style, high-performance deep
  learning library. In: Advances in Neural Information Processing Systems 32,
  pp. 8024--8035. Curran Associates, Inc. (2019),
  \url{http://papers.neurips.cc/paper/9015-pytorch-an-imperative-style-high-performance-deep-learning-library.pdf}

\bibitem{AlphaGo}
Silver, D., Huang, A., Maddison, C.J., Guez, A., Sifre, L., van~den Driessche,
  G., Schrittwieser, J., Antonoglou, I., Panneershelvam, V., Lanctot, M.,
  Dieleman, S., Grewe, D., Nham, J., Kalchbrenner, N., Sutskever, I.,
  Lillicrap, T., Leach, M., Kavukcuoglu, K., Graepel, T., Hassabis, D.:
  Mastering the game of {Go} with deep neural networks and tree search. Nature
  \textbf{529}(7587),  484--489 (Jan 2016). \doi{10.1038/nature16961}

\bibitem{Leaky}
Xu, B., Wang, N., Chen, T., Li, M.: {Empirical Evaluation of Rectified
  Activations in Convolutional Network} (Nov 2015),
  \url{http://arxiv.org/abs/1505.00853}

\bibitem{QPyTorch}
Zhang, T., Lin, Z., Yang, G., Sa, C.D.: Qpytorch: {A} low-precision arithmetic
  simulation framework. CoRR  \textbf{abs/1910.04540} (2019),
  \url{http://arxiv.org/abs/1910.04540}

\bibitem{DoReFa}
Zhou, S., Ni, Z., Zhou, X., Wen, H., Wu, Y., Zou, Y.: Dorefa-net: Training low
  bitwidth convolutional neural networks with low bitwidth gradients. CoRR
  \textbf{abs/1606.06160} (2016), \url{http://arxiv.org/abs/1606.06160}

\end{thebibliography}
\end{document}